\documentclass[10pt,twocolumn,letterpaper]{article}

\usepackage{iccv}              

\definecolor{DodgerBlue}{RGB}{29, 167, 239}

\definecolor{revBlue}{RGB}{0,0,0}
\newcommand{\cl}[1]{\textcolor{revBlue}{#1}}

\newcommand{\cmark}{\textcolor{green!80!black}{\ding{51}}}
\newcommand{\xmark}{\textcolor{red}{\ding{55}}}

\definecolor{iccvblue}{rgb}{0.21,0.49,0.74}
\usepackage[pagebackref,breaklinks,colorlinks,allcolors=iccvblue]{hyperref}
\usepackage{multirow}
\usepackage{pifont}
\usepackage{xcolor}

\newcommand*{\affaddr}[1]{#1} 
\newcommand*{\affmark}[1][*]{\textsuperscript{#1}}
\newcommand{\authormark}[2][]{%
  \begingroup
  \def\@thefnmark{#1}%
  \footnote{#2}%
  \endgroup
}

\title{EgoAgent: A Joint Predictive Agent Model in Egocentric Worlds \vspace{-2ex}}

\author{
Lu Chen\affmark[1]\textsuperscript{*}~~~~~ 
Yizhou Wang\affmark[2]\textsuperscript{*}~~~~~ 
Shixiang Tang\affmark[2]\textsuperscript{$\dag$}~~~~~ 
Qianhong Ma\affmark[3]~~~~~ 
Tong He\affmark[4] \\
Wanli Ouyang\affmark[2]~~~~~
Xiaowei Zhou\affmark[1]~~~~~
Hujun Bao\affmark[1]~~~~~
Sida Peng\affmark[1]\textsuperscript{$\dag$}\\
\affaddr{\affmark[1]State Key Lab of CAD\&CG, Zhejiang University~~~~~}
\affaddr{\affmark[2]The Chinese University of Hong Kong} \\
\affaddr{\affmark[3]Shanghai Jiao Tong University~~~~~}
\affaddr{\affmark[4]Shanghai Artificial Intelligence Laboratory} \\
\small
$^\ast$Equal contribution ~~~~~~~~$^\dag$Corresponding author
}

\begin{document}
\twocolumn[
\maketitle
\begin{center}
    \captionsetup{type=figure}
    \centering
    \vspace{-3ex}
    \includegraphics[width=0.93\linewidth]{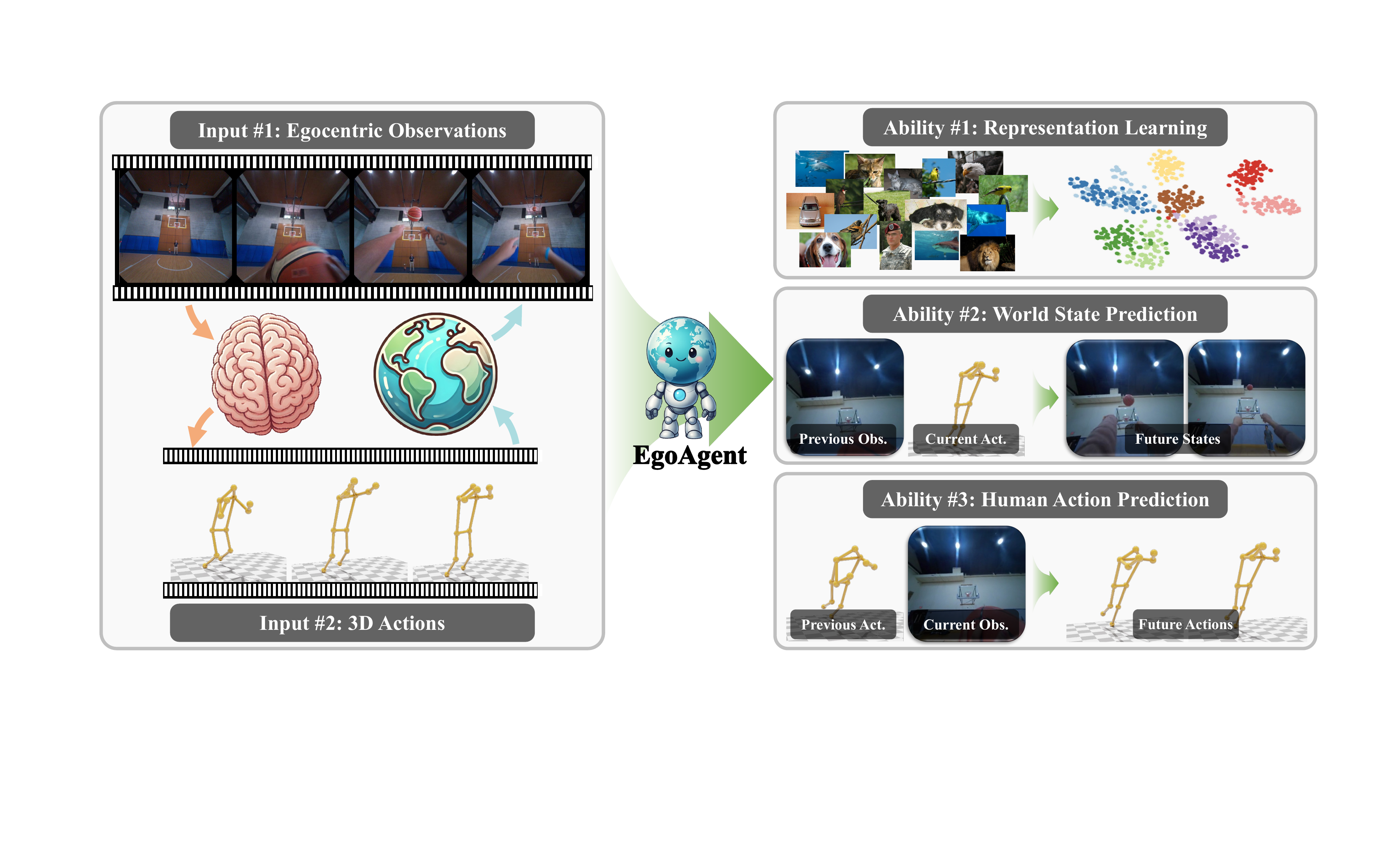}
    \vspace{-1ex}
    \captionof{figure}{(a) Humans learn from the world by continuously perceiving the egocentric world, predicting future states, and taking actions to achieve their goals. (b) Inspired by the human learning process, we propose \textbf{EgoAgent}, a joint predictive agent model in egocentric worlds that can learn to represent, predict, and act from egocentric videos and synchronized 3D human body motions.}
    \label{fig:teaser-image}
\end{center}
]
\begin{abstract}
Learning an agent model that behaves like humans---capable of jointly perceiving the environment, predicting the future, and taking actions from a first-person perspective---is a fundamental challenge in computer vision.
Existing methods typically train separate models for these abilities, which fail to capture their intrinsic relationships and prevent them from learning from each other.
Inspired by how humans learn through the perception-action loop, we propose {EgoAgent}, a unified agent model that simultaneously learns to represent, predict, and act within a single transformer.
EgoAgent explicitly models the causal and temporal dependencies among these abilities by formulating the task as an interleaved sequence of states and actions.
It further introduces a joint embedding–action–prediction architecture with temporally asymmetric predictor and observer branches, enabling synergistic optimization across all three capabilities.
Comprehensive evaluations of EgoAgent on representative tasks such as image classification, egocentric future state prediction, and 3D human motion prediction demonstrate the superiority of our method.
The code and trained models will be publicly available at \url{https://github.com/zju3dv/EgoAgent}. 
\end{abstract}    
\section{Introduction}
\label{sec:intro}

``\textit{Cognition is embodied; it arises from bodily interactions with the world,}" as stated by Linda B. Smith\footnote{\href{https://www.nasonline.org/directory-entry/linda-b-smith-v0ddqs/}{Linda Brawn Smith} is a cognitive scientist recognized for her work in developmental psychology and cognitive science, with over 53,000 citations. She won the David E. Rumelhart Prize for theoretical contributions to cognitive science and is a member of both the National Academy of Sciences and the American Academy of Arts and Science.} in~\cite{thelen2001dynamics}. 
This perspective emphasizes that cognition emerges from continuous, dynamic interactions between the brain, body, and environment, rather than relying on symbolic systems detached from real-world experiences~\cite{thelen1994dynamic,barsalou2008grounded,embodied-cognition}.
Through these ongoing interactions, humans naturally acquire foundational abilities such as \textit{visual perception}, \textit{predictive understanding of world dynamics}, and \textit{action anticipation}.
Replicating these abilities in artificial systems remains one major challenge in AI research, as highlighted by Yann LeCun~\cite{lecun2022path}.
Motivated by this human learning process, we propose an agent model that jointly learns these abilities from egocentric RGB observations and their responsive actions.
Such abilities are essential for applications in robotics, augmented reality, and gaming, where agents must understand what is happening in the world, anticipate what will happen next, and decide what actions to take. %

Existing approaches typically address these capabilities in isolated tasks: (1)
Visual representation learning~\cite{caron2021emerging,he2022masked,wang2022revisiting}, which encodes high-level representations of human observations of the world; (2)
World models~\cite{mendonca2023structured,yang2023learning}, which learn predictive representations of world state transitions conditioned on actions;
(3) Action prediction~\cite{martinez2017human,cao2020long}, which forecasts future human actions based on prior motion.
However, research in cognitive science, particularly the famous \emph{Common Coding Theory}~\cite{prinz1997perception,barsalou2008grounded}, claims that perception and action are not separate processes but deeply intertwined and mutually reinforcing, operating within a shared representational space.
Therefore, modeling these abilities within a unified framework is not only mutually beneficial but also represents a more principled approach for agent models to achieve human-like understanding in egocentric worlds.

While combining these abilities offers clear benefits, integrating them into one model remains challenging. Human interaction with the world involves a continuous loop of perceiving egocentric observations and taking 3D actions, creating complex interdependencies between observations and actions in time and causality. A successful model must therefore learn from this interleaved and tightly-coupled structure. The core difficulty, which remains an open question, lies in designing learning frameworks and supervision signals to capture these intricate connections.

In response, we propose \emph{EgoAgent}, a joint predictive agent model that simultaneously learns to represent world observations, predict future states, and act based on learned representations (Fig.~\ref{fig:teaser-image}). 
EgoAgent introduces two key innovations:
(1) Formulating human-world interactions as interleaved sequences of ``state-action-state-action'' tokens processed by the causal attention mechanism.
This design explicitly models both how observations trigger actions and how those actions, in turn, influence the subsequent states of the world.
(2) A joint embedding-action-prediction (JEAP) architecture with temporally asymmetric predictor-observer branches. The predictor leverages past information to forecast future human actions and world states, while the observer extracts the target world states from raw observations. By minimizing the prediction error between these two branches in a shared semantic feature space, JEAP is tailored to learn both world dynamics and human policy based on high-level visual representations, thereby integrating the three tasks into a cohesive learning framework.

{Extensive experiments show that EgoAgent effectively handles perception, prediction, and action tasks, even surpassing state-of-the-art methods across individual benchmarks. Specifically,
it outperforms the leading egocentric video pretraining method~\cite{venkataramanan2023imagenet} by \textbf{+1.32}\% Top1 accuracy on ImageNet-1K~\cite{deng2009imagenet} for image classification \cl{and \textbf{+3.90}\% success rate of moving cube in the TriFinger simulator~\cite{wuthrich2020trifinger}.}\footnote{Image classification serves as a standard benchmark for visual perception, while TriFinger evaluates the transferability of learned representations to egocentric manipulation with a three-finger robot. Together, these tasks assess a model's ability to learn meaningful visual representations.}
On Ego-Exo4D~\cite{grauman2024ego}, it achieves \textbf{+16.28}\% Top1 accuracy and \textbf{+16.95}\% mAP for future state prediction, while also exceeding 3D human motion prediction methods by improving MPJPE by \textbf{-0.82} at a 30 fps prediction rate.

Our contributions are as follows: (1) We develop the first agent model that can simultaneously learn to represent egocentric observations, predict future states, and generate 3D actions in a unified representational space.
(2) We propose a JEAP architecture with temporally asymmetric branches to learn both semantic understanding and temporal transitions of the dynamic multi-modal egocentric world.
(3) We evaluate our EgoAgent on egocentric video-action datasets and show its superior abilities in visual representation, world states prediction, and 3D human motion prediction tasks.

\section{Related Work}
\label{sec:related}

\begin{figure*}[th]
    \centering
    \includegraphics[width=0.98\linewidth]{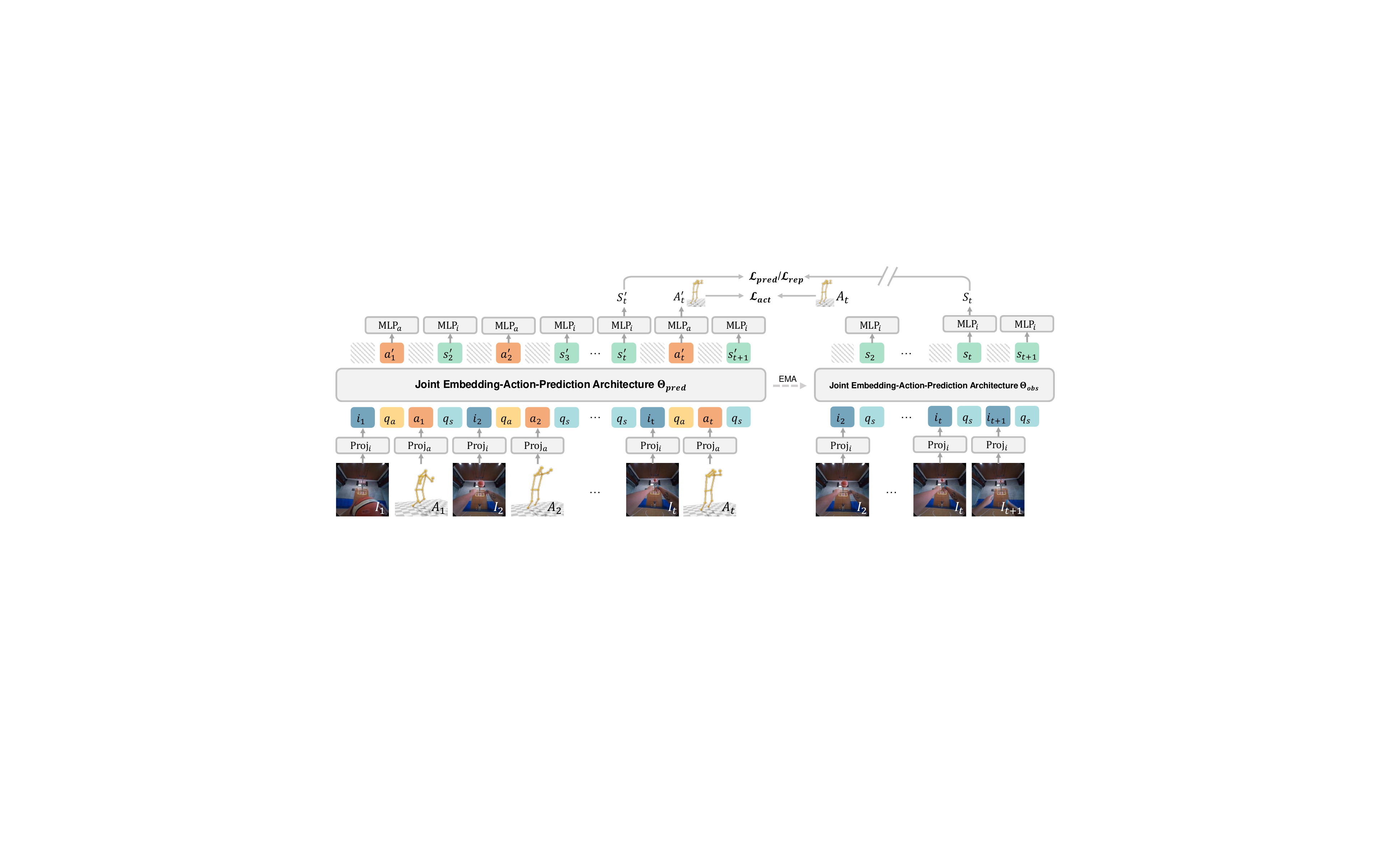}
    \vspace{-0.5em}
    \caption{The overall framework of EgoAgent. EgoAgent adopts the Joint Embedding-Action-Prediction Architecture to project input egocentric video frames $I_t$ and 3D human poses $A_t$ into image tokens $i_t$ and action tokens $a_t$. Learnable action query token $q_a$ and future state query token $q_s$ are appended after the image and action tokens to stimulate EgoAgent's predictor $\Theta_{pred}$ to predict the next action $A'_t$ and world state $S'_{t+1}$. The action prediction loss $\mathcal{L}_{act}$ is used to supervise the predicted actions. Since EgoAgent predicts world states $S'_{t+1}$ in a semantic feature space, the target world state $S_{t+1}$ is obtained from a momentum observer $\Theta_{obs}$. Features from the predictor and observer are aligned through a state prediction loss $\mathcal{L}_{pred}$. When 3D poses are not paired with video frames, EgoAgent randomly crops different views from the video and learns representative features only with image tokens $i_t$ and state query tokens $q_s$ by a self-supervised representation loss $\mathcal{L}_{rep}$. For easier understanding, we only illustrate the supervisions on predicting the state $S_t$ and action $A_t$.}
    \label{fig:framework}
    \vspace{-1em}
\end{figure*}

\noindent\textbf{World Models and Agent Models.}
Ha and Schmidhuber~\cite{ha2018world} defined the world model as a predictive representation of the environment, while the agent model utilizes this learned representation to make decisions and take actions.
Yann LeCun~\cite{lecun2022path} further envisioned that an agent model can generate actions either reactively or through iterative optimization using the world model.
Recently, \emph{generative world models}~\cite{yang2023learning,alonso2024diffusion,xiang2024pandora} have demonstrated success in autonomous driving~\cite{wang2024driving}, robotics~\cite{ha2018recurrent}, and game control~\cite{bamford2020neural}, mostly employing an encoder-decoder architecture to learn transitions between video frames.
GAIA-1~\cite{hu2023gaia} maps multi-modal driving signals into discrete tokens, and processes them using an autoregressive transformer.
In contrast, \emph{JEPA world models}~\cite{assran2023self,bardes2024revisiting,garrido2024learning} focus on learning state transitions through the joint embedding predictive architecture (JEPA)~\cite{lecun2022path} without generating videos explicitly.
MC-JEPA~\cite{bardes2023mc} incorporates self-supervised learning with optical flow estimation to learn content features and motion dynamics.
In this work, we extend the JEPA world model into a reactive agent model, which not only predicts world states based on human actions but also predicts actions based on learned world state transitions.

\noindent\textbf{Egocentric Visual Representation Learning.}
Traditional visual representation learning methods are trained on large-scale image datasets, such as ImageNet~\cite{deng2009imagenet}, using either supervised~\cite{he2016deep,khosla2020supervised,wang2022revisiting} or self-supervised methods~\cite{he2020momentum,caron2021emerging,chen2020simple}. 
However, these datasets primarily consist of third-person perspectives, which do not fully align with how humans perceive the world. To bridge this gap, researchers have turned to egocentric videos, which offer a more natural and immersive viewpoint for learning visual representations.
Models like R3M~\cite{nair2022r3m} and VIP~\cite{ma2022vip} leverage the temporal relationships in egocentric videos to learn generalizable representations from Ego4D~\cite{grauman2022ego4d} for robotics applications.
Recently, DoRA~\cite{venkataramanan2023imagenet} learns object-level representations by tracking movements on an egocentric long video dataset called WalkingTours.
In this paper, we propose an agent model that not only learns robust representations from egocentric videos but also predicts future states and actions.

\noindent\textbf{3D Human Motion Prediction.}
Human motion prediction models~\cite{yuan2019ego,liu2021aggregated,wang2021pvred} aim to forecast future body movements given historical poses.
Aksan et al.~\cite{aksan2021spatio} proposed a transformer-based architecture 
to capture the spatial-temporal dependencies of human motion over short and long horizons. Cui et al.~\cite{cui2020learning} represent the human skeleton as a dynamic graph to adaptively learn joint connection strengths for better prediction accuracy.
Other works~\cite{maeda2022motionaug,zhang2024incorporating} incorporate physics-based priors to ensure the generated motions adhere to physical principles. 
Recent research has integrated multimodal cues such as textual instructions~\cite{mao2022weakly}, eye gaze~\cite{zheng2022gimo}, and 3D objects~\cite{yan2024forecasting} as conditions for human motion prediction.
Notably, Cao et al.~\cite{cao2020long} proposed a three-stage framework that consists of goal prediction, path planning, and pose finalization, 
which effectively enhances scene-aware motion prediction.
\section{Method}
\label{sec:method}

Our goal is to develop an agent model that can \textit{represent world states}, \textit{predict future states}, and \textit{generate future actions} by simulating how humans interact with the world through egocentric vision and 3D skeletal motion.
The overview of our framework is presented in Fig.~\ref{fig:framework}.
We first introduce our joint embedding-action-prediction architecture in Sec.~\ref{sec:joint embedding-action predictive world model}. We then introduce how EgoAgent can be trained in Sec.~\ref{sec:learning_world_model}.
Finally, we demonstrate the application of EgoAgent in the three tasks in Sec.~\ref{sec:applications}.

\subsection{Joint Embedding-Action-Prediction Architecture for EgoAgent}
\label{sec:joint embedding-action predictive world model}

Given egocentric observations $V=\{I_1, I_2, ...I_{t}\}$ and human actions $A=\{A_1, A_2, ...A_{t}\}$,
our agent model aims to predict the future world state $S_{t+1}$ associated with the next observation $I_{t+1}$ while simultaneously generate plausible human action $A_{t+1}$.
The world states are represented as semantic embeddings, whereas human actions are 3D poses.

\noindent{\textbf{Interleaved joint prediction.}
EgoAgent begins by encoding the egocentric video $V$ and human action sequence $A$ into semantic embeddings.
For video frames, we follow the approach of ViT~\cite{dosovitskiy2020image}, applying a convolutional layer to each video frame $I_t$ to produce a feature map.
This map is then subdivided and flattened to obtain the image feature $i_t$.
For human actions, we utilize a convolutional layer followed by layer normalization (LN) and a Gaussian error linear unit (GeLU) activation to map the 3D body pose $A_t$ into action feature $a_t$.
These feature vectors are then processed by a causal transformer network.

As illustrated in Fig.~\ref{fig:framework}, at each time step $t$, we assemble a structured sequence of interleaved tokens, including the image token $i_{t}$, an action query token $q_{a}$, the action token $a_{t}$, and a future state query token $q_{s}$. 
The query tokens are learnable embeddings that serve as explicit prompts, directing the model to predict either the next action or a future world state.
Notably, we insert $q_{a}$ right after the image tokens and $q_{s}$ after the action tokens to model the causal and temporally interlaced nature of perceiving observations and taking actions.
Specifically, enabled by transformer's causal attention mechanism, $q_{a}$ incorporates all preceding image tokens $i_{[0:t]}$ and action tokens $a_{[0:t-1]}$ to predict $a'_{t}$.
Similarly, $q_{s}$ integrates $i_{[0:t]}$ and the newly appended $a_{[0:t]}$ to predict $s'_{t+1}$. 
We utilize separate MLPs to map these embeddings to 3D actions $A'_{t}$ and world states $S'_{t+1}$.

\noindent{\textbf{JEPA with temporally asymmetric branches.}
Unlike actions, which can be directly obtained and supervised, world states must be learned from observations while also be used as supervision signals for state prediction.
To achieve this, we propose a temporally asymmetric predictor-observer architecture based on JEPA~\cite{lecun2022path,garrido2024learning}, a framework that combines representation learning with predictive learning.
In addition to the predictor branch executing the interleaved joint prediction, we introduce an observer branch that only processes image inputs by (1) extracting current-frame features for self-supervised representation learning and (2) providing next-frame features for future state supervision alongside the predictor.
Specifically, the observer branch takes image tokens $i_{t}$ and state query tokens $q_s$ as input, and outputs current state embeddings $s_t$, which are then fed into the same MLP for obtaining world states $S_{t}$ from image embeddings in the predictor.}
A key advantage of our query-based design is that it effectively decouples the shared state/representation components from the predictor's action-specific ones.
This prevents conflicting gradient updates, allowing the observer's parameters to be updated as an exponential moving average (EMA) of the predictor's, despite the observer not processing the action modality.


\noindent{\textbf{Learning in semantic feature space.}
Notably, unlike previous methods~\cite{bai2024sequential,yang2023learning} that rely on pretrained reconstruction-based tokenizers, such as VQGAN~\cite{esser2021taming}, to convert images into discrete tokens, we employ learnable convolutional layers to project images into continuous semantic embeddings.
Discrete tokens prioritize pixel-level reconstruction over high-level semantic information, whereas humans make predictions based on abstract concepts rather than pixels. Therefore, our approach aligns more closely with human reasoning and significantly enhances the model's performance in future state prediction (see Sec.~\ref{sec:exp-world-prediction}).
}

\subsection{{Training EgoAgent}}
\label{sec:learning_world_model}

\noindent\textbf{Base model.}
We employ InternLM, an open-source LLM~\cite{team2023internlm}, as the foundational architecture for EgoAgent, adapting it to the JEAP architecture mentioned in Sec.~\ref{sec:joint embedding-action predictive world model}.
This is motivated by two key considerations:
(1) Our objective of near-future world state and action prediction naturally aligns with the next-token prediction mechanism inherent in LLMs.
(2) The structured yet flexible architecture of LLMs provides a scalable foundation for our specific tasks.
We do NOT initialize the LLM with any pretrained weights, as we consider that visual perception and prediction abilities can be learned without language priors~\cite{bai2024sequential}.

\noindent{\textbf{Supervisions for predicting world states and actions.}}
EgoAgent consists of two branches: a predictor branch, which predicts future world states $S_{t+1}^{'}$ and actions $A_{t}^{'}$, and an asymmetric observer branch, which extracts the future world states $S_{t+1}$ using only the egocentric image $I_{t+1}$.
Given the input token sequence at time step $t$, the loss functions for future state and action prediction are defined as:
    \begin{align}
    \small
       \mathcal{L}_{act}(t)&=\mathcal{L}_1(A_t',A_t), \\
    \mathcal{L}_{pred}(t)&=\mathcal{L}_{dino}(S_{t+1}', sg[S_{t+1}]),       
    \end{align}
where $sg[\cdot]$ denotes the stop-gradient operation, $\mathcal{L}_1$ and $\mathcal{L}_{dino}$ represent the L1 loss and the DINO loss~\cite{caron2021emerging}, respectively.
Following common practices in self-supervised learning~\cite{caron2021emerging, he2020momentum}, sg[$\cdot$] is applied to block the gradients from back-propagating to the observer branch.
The weights of the observer branch, including the MLPs, are updated at each training iteration using an EMA of the predictor's.

\noindent{\textbf{Self-supervision for learning powerful representations.}}
When humans learn to interact with the environment, they first develop an understanding of the observed world and objects within, which then aids them in predicting future states of the world and making appropriate responses. Inspired by this process, we introduce an additional self-supervised learning loss on EgoAgent to facilitate learning representative features from egocentric videos from scratch:
\begin{align}
\small
    \mathcal{L}_{rep}(t)&=\mathcal{L}_{dino}(\Theta_{pred}(I_t^{v1})
    , \Theta_{obs}(I_t^{v2})).  
\end{align}
Here, $\Theta_{pred}$ and $\Theta_{obs}$ denote the predictor and observer networks of EgoAgent, respectively, $I_t^{v1}$ and $I_t^{v2}$ denote the two different views derived from the egocentric image $I_t$.

\noindent\textbf{Overall objective function.}
Given an input interleaved token sequence containing $t$ time steps, we have the overall objective function $\mathcal{L}$ defined as:
\begin{equation}
\small
    \mathcal{L} = \frac{1}{t}\sum_{k=0}^t{(
    \lambda_{rep}\mathcal{L}_{rep} + \lambda_{pred}\mathcal{L}_{pred} +
    \lambda_{act}\mathcal{L}_{act}),
    }
\end{equation}
where $\lambda_{rep}$, $\lambda_{pred}$, and $\lambda_{act}$ are the corresponding loss weights for representing the world, predicting future world states and predicting actions, respectively.

\subsection{{Task-Specific Inference}}
\label{sec:applications}
After training, EgoAgent inherently acquires the abilities to solve the three tasks designed to mirror human cognition:

\noindent\textbf{World state prediction.} Given a sequence of egocentric video frames and human motion, EgoAgent is capable of predicting the future world state $S_{t+1}$ by
$
S_{t+1} = \Theta(I_1, A_1, ..., I_{t}, A_t),
$
where $\Theta$ denotes the EgoAgent model.
Once predicted, the future world image can then be retrieved by measuring similarity within the learned feature space, enabling applications that require foresight into upcoming visual states.

\noindent\textbf{Human action prediction.} From an egocentric view and informed by previous body motions, EgoAgent can generate plausible future action $A_t$ represented as 3D human motion by
$
A_t =\Theta(I_1, A_1, ..., A_{t-1}, I_{t}).
$
This predictive capability of human motion is essential for applications in humanoid robotics, virtual reality, and interactive gaming, where anticipating human-like movements is crucial.

\noindent{\textbf{Visual representation.}} EgoAgent learns world states as feature representations, allowing it to perceive meaningful semantic features from input images
by $ S_t = \Theta(I_t).$
These features can be directly applied to representation tasks such as image classification and visuomotor policy learning.
\begin{table*}[htbp]
\centering
\caption{{Performance comparison across three tasks. (1) For world state prediction, we report the feature retrieval accuracy of the predicted next-frame features by the averaged Top1 accuracy and mAP on total timesteps $T=4$. (2) For human action prediction, we report the 3D motion prediction results on Ego-Exo4D by MPJPE (cm) and MPJVE (cm/s) with time gaps of 1/30 and 1/10 seconds. (3) For visual representation, we report the $k$-NN evaluation results on ImageNet-100 and ImageNet-1K by Top1 and Top5 accuracy (\%). We evaluate with the officially released checkpoints of representation models trained on egocentric video datasets and train all motion prediction models from scratch. $^*$ denotes the model trained by ourselves for a fair comparison, $\downarrow$ indicates lower is better.}}
\vspace{-0.5em}
\resizebox{\linewidth}{!}{
\begin{tabular}{llcccccccccc}
\toprule
\multicolumn{1}{l}{\multirow{3}{*}{Method}} &
  \multicolumn{1}{l}{\multirow{3}{*}{Training Dataset}} &
  \multicolumn{2}{c}{\textbf{World State Prediction}} &
  \multicolumn{4}{c}{\textbf{3D Human Motion Prediction}} &
  \multicolumn{4}{c}{\textbf{Visual Representation}} \\ \cmidrule(r){3-4} \cmidrule(r){5-8} \cmidrule(r){9-12}
\multicolumn{1}{c}{} &
  \multicolumn{1}{c}{} &
  \multicolumn{2}{c}{Ego-Exo4D} &
  \multicolumn{2}{c}{Ego-Exo4D (30fps)} &
  \multicolumn{2}{c}{Ego-Exo4D (10fps)} &
  \multicolumn{2}{c}{ImageNet-100} &
  \multicolumn{2}{c}{ImageNet-1K} \\
\multicolumn{1}{c}{} & \multicolumn{1}{c}{} & Top1 Acc. & mAP   & MPJPE$\downarrow$ & MPJVE$\downarrow$  & MPJPE$\downarrow$ & MPJVE$\downarrow$  & Top1  & Top5  & Top1  & Top5  \\ \midrule
VIP~\cite{ma2022vip}                  & Ego4D                & 1.67      & 6.10  & \xmark     & \xmark      & \xmark     & \xmark      & 8.04  & 21.14 & 1.59  & 1.59  \\
R3M~\cite{nair2022r3m}                  & Ego4D                & 24.42     & 37.26 & \xmark     & \xmark      & \xmark     & \xmark      & 4.82  & 14.42 & 0.77  & 0.77  \\
DINO~\cite{caron2021emerging}                 & WT$_\text{Venice}$                 & 28.24     & 43.42 & \xmark     & \xmark      & \xmark     & \xmark      & 40.12     & 63.74     & 22.18 & 35.85     \\
DoRA~\cite{venkataramanan2023imagenet}                 & WT                   & 30.15     & 45.01 & \xmark     & \xmark      & \xmark     & \xmark      & 55.08 & 78.06 & 34.52 & 52.50 \\
DoRA$^*$~\cite{venkataramanan2023imagenet}                & WT+Ego-Exo4D          & 29.78     & 44.34 & \xmark     & \xmark      & \xmark     & \xmark      & 51.74 & 74.14 & 31.46 & 47.82 \\ \midrule
Diffusion Policy-C~\cite{chi2023diffusion}   & Ego-Exo4D            & \xmark         & \xmark     & 28.07 & 206.96 & 27.95 & 114.75 & \xmark     & \xmark     & \xmark     & \xmark     \\
Diffusion Policy-T~\cite{chi2023diffusion}   & Ego-Exo4D            & \xmark         & \xmark     & 25.92 & 353.24 & 25.85 & 148.82 & \xmark     & \xmark     & \xmark     & \xmark     \\
HumanMAC~\cite{chen2023humanmac}             & Ego-Exo4D            & \xmark         & \xmark     & 19.21 & 94.22  & 17.68 & 77.43  & \xmark     & \xmark     & \xmark     & \xmark     \\
siMLPe~\cite{guo2023back}               & Ego-Exo4D            & \xmark         & \xmark     & 13.33 & 81.94  & 12.20 & 60.65  & \xmark     & \xmark     & \xmark     & \xmark     \\ \midrule
EgoAgent-300M        & WT+Ego-Exo4D          & 43.01     & 58.06 & 12.92 & 82.18  & 11.89 & 59.96  & 55.14 & 76.56 & 34.65 & 51.42 \\
EgoAgent-1B          & WT+Ego-Exo4D          & \textbf{46.43}     & \textbf{61.96} & \textbf{12.51} & \textbf{81.45}  & \textbf{11.65} & \textbf{58.99}  & \textbf{56.48} & \textbf{78.12} & \textbf{35.84} & \textbf{53.03} \\ \bottomrule
\end{tabular}
}
\label{tab:comparison}
\vspace{-1em}
\end{table*}

\section{Experiments}
\label{sec:exp}

\subsection{{Experimental Setups}}
\noindent{\textbf{Datasets.}}
We train EgoAgent on two egocentric datasets, WalkingTours (WT)~\cite{venkataramanan2023imagenet} and Ego-Exo4D~\cite{grauman2024ego}. 
WT is a video-only dataset with approximately 1.5M high-resolution frames.
Ego-Exo4D has 221.26 hours of egocentric videos
along with 376K manually labeled and 9.2M automatically generated 3D body poses. 
To eliminate the camera projection difference between Ego-Exo4D and WT, we undistort the raw fisheye sensor data in Ego-Exo4D into a pinhole camera model.
We apply a sliding window filter of 20 frames to extract continuous pose sequences, resulting in 1,410,119 clips of synchronized egocentric videos and 3D body poses. 
Based on the EgoPose split in Ego-Exo4D, we reserve the validation set of Ego-Exo4D-v2\footnote{The whole Ego-Exo4D dataset is divided into Ego-Exo4D-v1 and Ego-Exo4D-v2 on the official website (https://ego-exo4d-data.org).} for evaluation and use the remaining clips for training, yielding 1,378,672 training clips and 31,447 evaluation clips.


\noindent\textbf{Implementation Details.}
To alleviate the training burden, we sample one image frame every five frames while retaining all 3D body poses. Specifically, for each clip with 20 frames, we divide it into $T=4$ time steps and sample the first image frame at each step. This results in every time step consisting of one image frame paired with five frames of 3D body poses. 
For self-supervised learning on egocentric videos, 
We follow~\cite{venkataramanan2023imagenet} to avoid cropping noisy positive pairs and adopt two global crops, six local crops, and the same augmentations consistent with DINO~\cite{caron2021emerging}.

Following the configurations in~\cite{guo2024data}, we train EgoAgent with two model sizes, 300M and 1B, 
on the processed dataset for 72,000 iterations with a linear warm-up of 1800 iterations. 
We set the base learning rate to $6\times 10^{-4}$ with a cosine decay scheduler. The loss weights are set as $\lambda_{rep}=2, \lambda_{pred}=1,  \lambda_{act}=3$, respectively. All models are trained from scratch using FP16 to speed up. The training takes 25/60 hours in total with a batch size of 1920 on 32/48 NVIDIA A100 GPUs for the 300M/1B model, respectively.

\begin{figure}[t]
    \centering
    \includegraphics[width=\linewidth]{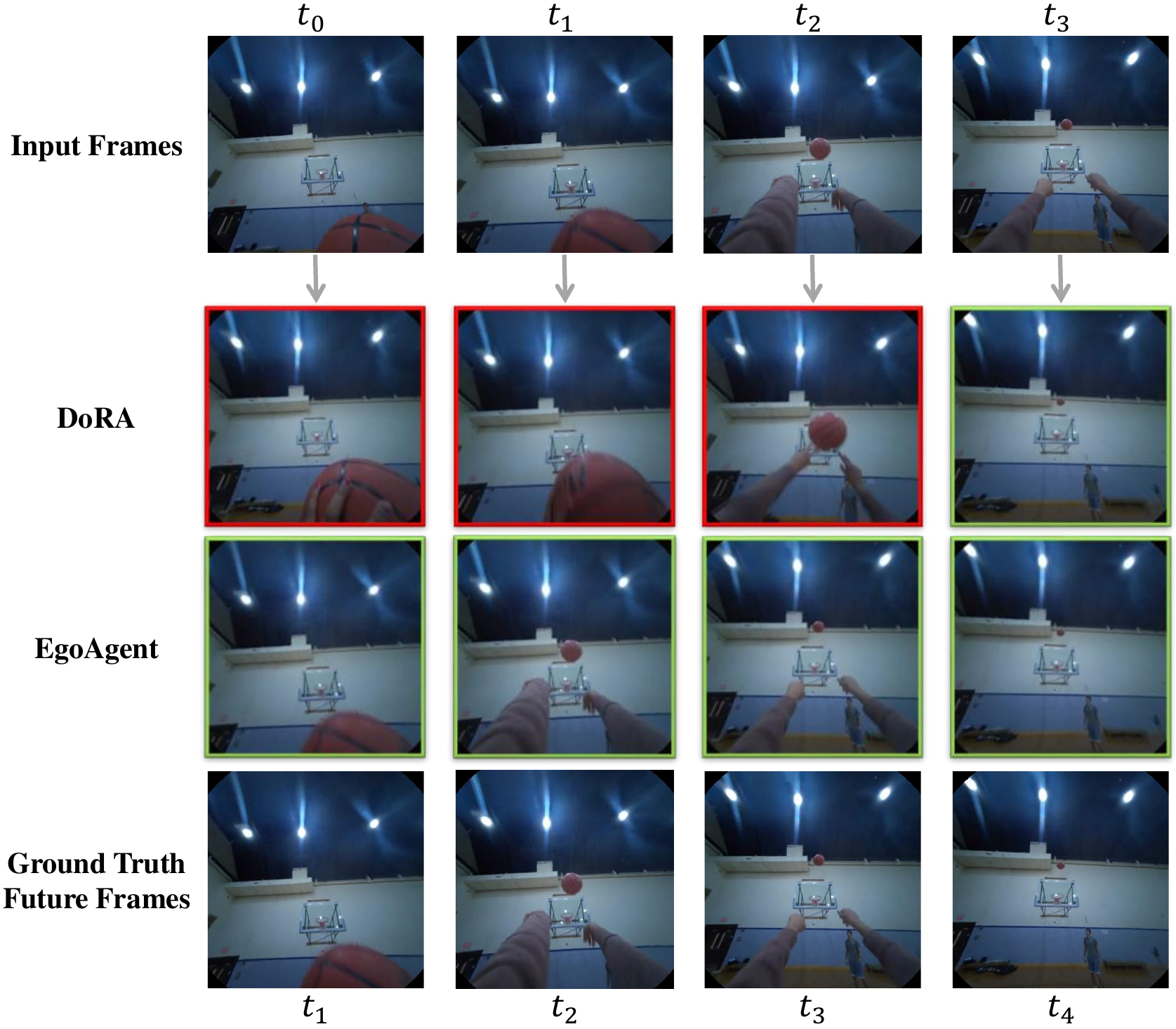}
    \vspace{-2em}
    \caption{Retrieval results for egocentric world state prediction. Images correctly retrieved are highlighted with green borders.}
    \label{fig:pred}
    \vspace{-1em}
\end{figure}

\subsection{World State Prediction}
\label{sec:exp-world-prediction}
To evaluate the effectiveness of EgoAgent in predicting future states of the world, we perform an egocentric video prediction task with feature retrieval. 
Specifically, given the egocentric video frames and 3D human poses at a certain time step, we build a query set that stores the EgoAgent-predicted features of the next frame and a gallery set that contains the features directly extracted from all video frames. 
At time step $t$, if the predicted world state $S'_{t+1}$ in the query set can correctly retrieve the corresponding $S_{t+1}$ in the gallery set, we treat it as a successful prediction.
For traditional representation models, since there are no action  conditions, the output state $S_{t+1}'$ would be exactly the state at time step $t$, \emph{i.e.}, $S_{t}$. To avoid retrieving this exact same state, when querying with input image $I_t$, we ignore the scene state $S_{t}$ in the gallery set.
Note that $S_{t}$ is NOT removed when testing EgoAgent.
Following common practice in retrieval tasks, we use Top1 accuracy and mean average precision (mAP) as metrics. 

As shown in Tab.~\ref{tab:comparison}, EgoAgent outperforms traditional representation models that trained with time-contrastive loss~\cite{ma2022vip,nair2022r3m,venkataramanan2023imagenet} or view-contrastive loss~\cite{caron2021emerging} on egocentric videos by a large margin. Specifically, EgoAgent-300M outperforms DoRA by \textbf{+12.86}\% in Top1 accuracy and \textbf{+13.05}\% in mAP, indicating that its predicted features are beyond the input images. In contrast, traditional models rely solely on semantic similarity, struggling to distinguish similar frames in a video. Furthermore, scaling up the model enhances performance, as EgoAgent-1B achieves further performance gains upon EgoAgent-300M with \textbf{+3.42}\% Top1 accuracy and \textbf{+3.90}\% mAP.
Note that when trained on our combined dataset, DoRA exhibits performance degradation in both world state prediction and visual representation tasks. We attribute this to its tracker's inability to track abrupt object motions in Ego-Exo4D. For further analysis, please refer to Sec.~\ref{sec:visual_represent}.
Fig.~\ref{fig:pred} visualizes the retrieval results of predicted features, where DoRA fails to retrieve the correct future states when humans have large movements, while EgoAgent successfully predicts the future features that retrieve the correct next-frame images.

\begin{figure}[t]
    \centering
    \includegraphics[width=\linewidth]{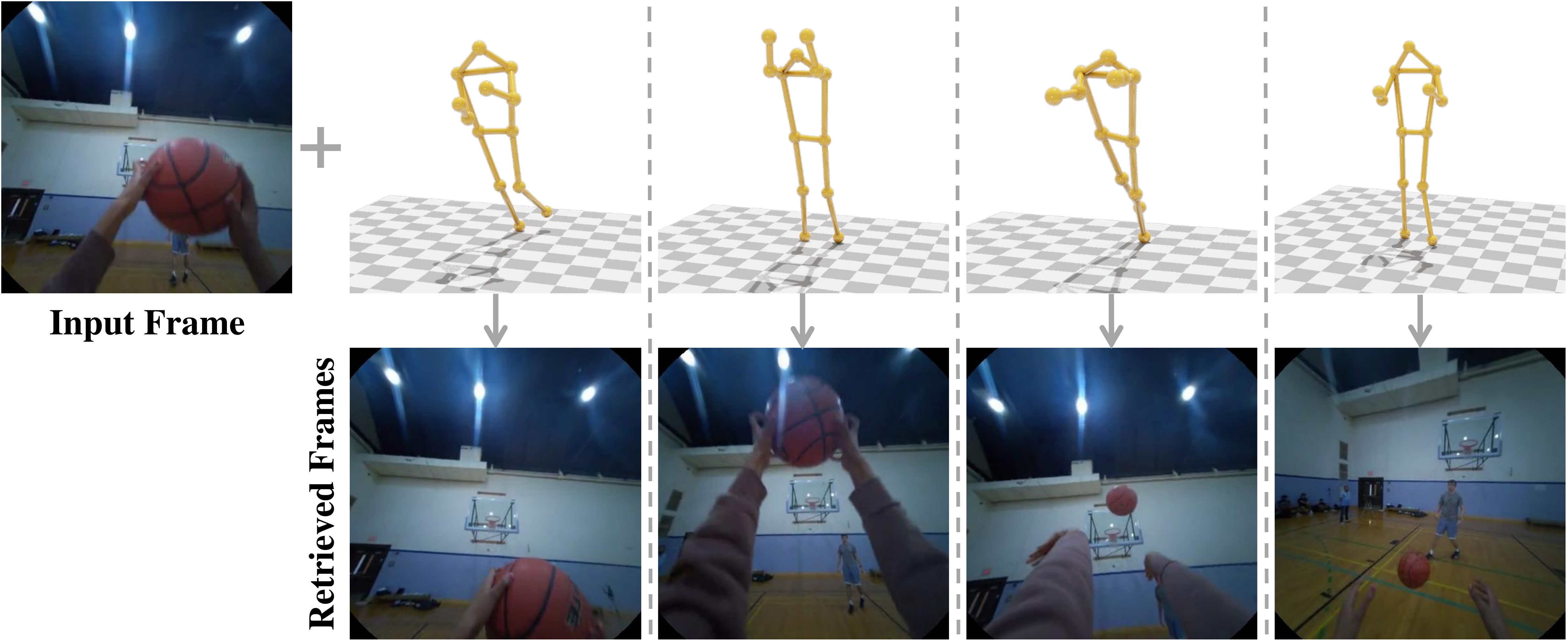}
    \vspace{-1.5em}
    \caption{Retrieval results for egocentric world state prediction from the same input observation with diverse human poses.}
    \label{fig:case_study}
    \vspace{-1.5em}
\end{figure}

To further evaluate whether EgoAgent accurately models the causal relationship between actions and future world states, we conduct a qualitative analysis by conditioning next-state predictions on various poses with the same initial observation, including dribbling, raising, shooting, and passing the ball. Fig.~\ref{fig:case_study} shows the retrieved images, where all predicted features successfully retrieve images that accurately reflect the intended motion dynamics. This demonstrates EgoAgent’s ability to understand both the semantics of actions and their consequences in the egocentric world.

\begin{figure*}[th]
    \centering
\includegraphics[width=.9\linewidth]{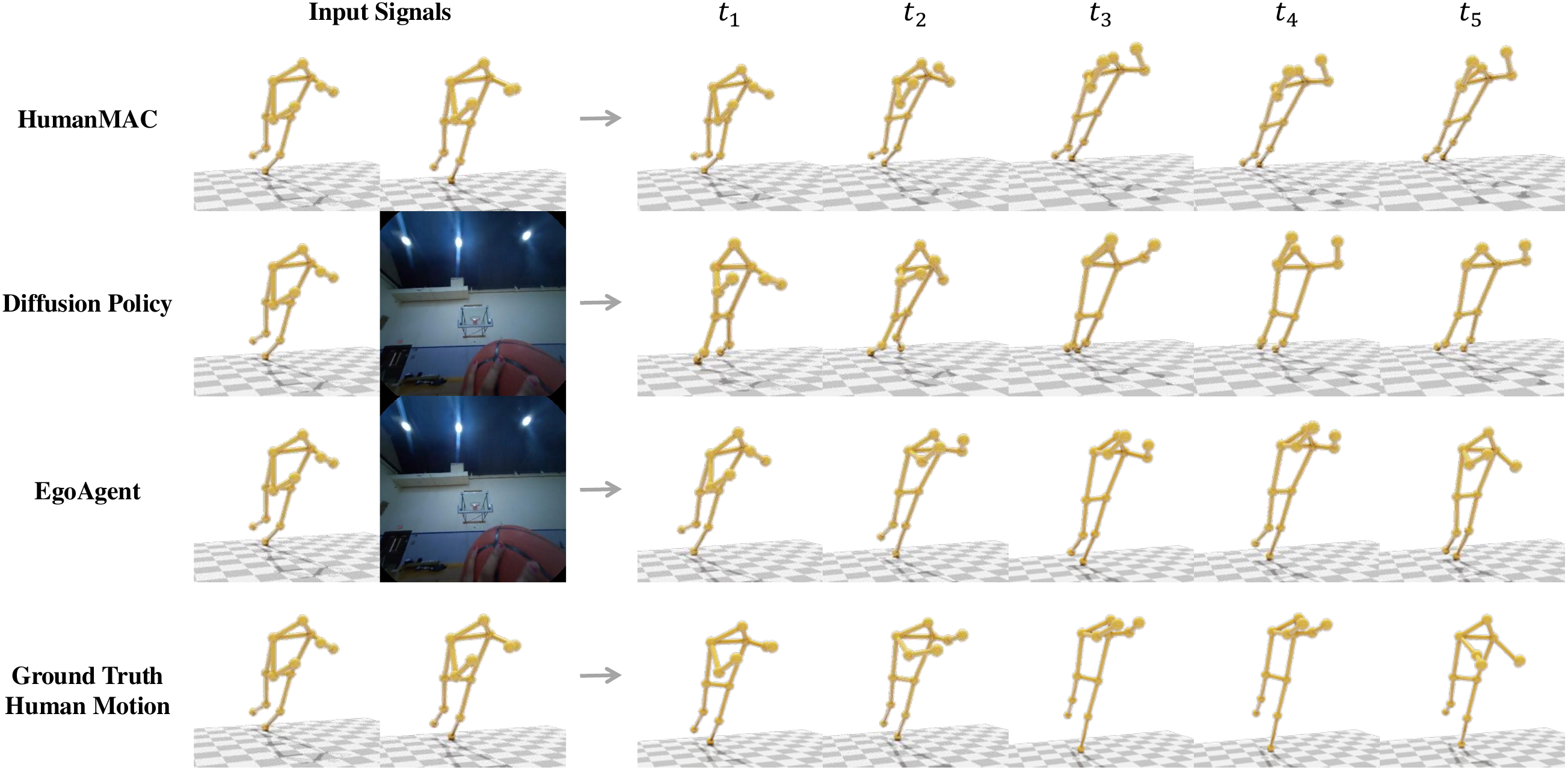}
\vspace{-0.5em}
    \caption{Visualizations of the 3D human motion prediction task. After observing the egocentric frames, EgoAgent can generate accurate human skeletons even though most of the body joints are not visible in the input frame.}
    \label{fig:3D motion}
    \vspace{-1em}
\end{figure*}

\subsection{3D Human Motion Prediction}
We compare EgoAgent with the state-of-the-art video-based motion generation model, Diffusion Policy~\cite{chi2023diffusion} with convolution (C) and transformer (T) architectures, and unconditional motion prediction models~\cite{guo2023back,chen2023humanmac} using the EgoPose from Ego-Exo4D~\cite{grauman2024ego}. 
Each testing clip contains 5 frames of egocentric images and 20 frames of 3D human motion. 
We set the prediction target as the last 15 frames of 3D motion. The images and first 5 frames of 3D motion are adopted as input for Diffusion Policy and EgoAgent, while only the first 5 poses are used as input for unconditional motion prediction models.
Following Ego-Exo4D~\cite{grauman2024ego}, the mean per-joint position error (MPJPE) in centimeters (cm) and the mean per-joint velocity error (MPJVE) in centimeters per second (cm/s) are adopted as evaluation criteria.

Quantitatively, as shown in Tab.~\ref{tab:comparison}, EgoAgent achieves state-of-the-art performance with the lowest MPJPE and competitive MPJVE. Specifically, EgoAgent-300M improves siMLPe~\cite{guo2023back} by \textbf{-0.41} and \textbf{-0.31} MPJPE on the evaluation set of 30 fps and 10 fps, respectively. When scaling up to 1B, the prediction errors are further reduced. These results demonstrate that EgoAgent can predict future actions accurately given egocentric observations and past actions.

We also evaluate the motion prediction results qualitatively. As illustrated in Fig.~\ref{fig:3D motion}, EgoAgent produces relatively small errors compared to the ground truth. Compared to Diffusion Policy~\cite{chi2023diffusion}, EgoAgent achieves higher accuracy on non-visible joints. Specifically, in the early frames ($t_1$ and $t_2$), the missing right hand in the image leads to large prediction errors in Diffusion Policy, whereas EgoAgent accurately estimates hand positions by leveraging prior motion cues. Compared to HumanMAC~\cite{chen2023humanmac}, EgoAgent shows smaller accelerated errors. The predicted skeletons in HumanMAC gradually lean toward the floor ($t_{[3:5]}$) after several frames, whereas EgoAgent integrates visible body part information from egocentric observations to correct accelerated errors, leading to more stable motion predictions.

\subsection{Visual Representation} \label{sec:visual_represent}
We examine EgoAgent's visual perception ability by two aspects: (1) image classification accuracy by $k$-NN on ImageNet-100 and ImageNet-1K~\cite{deng2009imagenet}, which is the widely-used golden standard for evaluating visual representations, and (2) egocentric manipulation success rate in an embodied simulator called TriFinger~\cite{wuthrich2020trifinger}, which measures the transferability of learned visual representations to visual-motor control. For both tasks, the models are frozen and used to extract visual features. For image classification, we utilize a $k$-nearest neighbor classifier with $k=20$.

As shown in Tab.~\ref{tab:comparison}, EgoAgent achieves the best representation performance. Specifically, EgoAgent-1B outperforms the officially released DoRA by \textbf{+1.40}\% and \textbf{+1.32}\% Top1 accuracy on ImageNet-100 and ImageNet-1K, respectively, demonstrating that learning to predict world states and actions can facilitate the task of representing the world. 
{For a fair comparison, we train DoRA on the same combined dataset (WT and Ego-Exo4D), which instead results in worse performance than training on WT alone. We suggest this is because, unlike WT videos, which feature stable motion, Ego-Exo4D includes fewer objects but more dynamic movements (objects frequently move out of sight). As a result, the tracker in DoRA struggles to maintain consistent object tracking, leading to degraded performance. In contrast, EgoAgent learns robust representation by modeling the temporal causality of state features and actions.}

\begin{table}[tb]
\centering
\caption{{Success rate (\%) on the TriFinger~\cite{wuthrich2020trifinger} benchmark, where each model's pretrained visual representation is fixed, and additional linear layers are trained as the policy network.}}
\vspace{-0.5em}
\label{tab:trifinger}
\resizebox{\linewidth}{!}{%
\begin{tabular}{llcc}
\toprule
Methods       & Training Dataset & Reach Cube & Move Cube \\
\midrule
DINO~\cite{caron2021emerging}         & WT$_\text{Venice}$       & 78.03     & 47.42     \\
DoRA~\cite{venkataramanan2023imagenet}          & WT        & 82.40     & 48.13     \\
DoRA~\cite{venkataramanan2023imagenet}          & WT+Ego-Exo4D           & 81.55     & 45.41     \\
\midrule
EgoAgent-300M & WT+Ego-Exo4D      & 82.61    & 54.21      \\
EgoAgent-1B   & WT+Ego-Exo4D      & \textbf{85.72}      & \textbf{57.66}   \\
\bottomrule
\end{tabular}%
}
\vspace{-1em}
\end{table}
\begin{table*}[thp]
  \centering
  \caption{{Ablation studies: removing one of the three tasks during training (b-d), training on a single task (e-g), and using pretrained image latents from VQGAN’s reconstruction-based latent space (h-i). All variants are trained with 14,400 iterations.}}
  \vspace{-0.5em}
  \resizebox{.86\textwidth}{!}{
    \begin{tabular}{lccccccccc}
    \toprule
    \multirow{4}[0]{*}{Method} & \multicolumn{3}{c}{\multirow{2}[0]{*}{Training Task}} & \multicolumn{2}{c}{\textbf{World State Prediction}} & \multicolumn{2}{c}{\textbf{Human Motion Prediction}} & \multicolumn{2}{c}{\textbf{Visual Representation}} \\ \cmidrule(r){5-6} \cmidrule(r){7-8} \cmidrule(r){9-10}
    & \multicolumn{3}{c}{} & \multicolumn{2}{c}{Ego-Exo4D} & \multicolumn{2}{c}{Ego-Exo4D (@30fps)} & \multicolumn{1}{c}{ImgNet-100} & \multicolumn{1}{c}{ImgNet-1K} \\ \cmidrule(r){2-4} \cmidrule(r){5-6} \cmidrule(r){7-8} \cmidrule(r){9-10}
    & \multicolumn{1}{c}{Pred.} & \multicolumn{1}{c}{Act.} & \multicolumn{1}{c}{Rep.} & \multicolumn{1}{c}{Top1 Acc.} & \multicolumn{1}{c}{mAP} & \multicolumn{1}{c}{MPJPE$\downarrow$} & \multicolumn{1}{c}{MPJVE$\downarrow$} & \multicolumn{1}{c}{Top1 Acc.} & \multicolumn{1}{c}{Top1 Acc.} \\
    \midrule
    (a) Baseline &   \cmark    & \cmark      &   \cmark   & \textbf{37.77}      &  \textbf{53.72}     & 14.49 & 88.61 & \textbf{41.64} & \textbf{22.28}  \\
    \midrule
    (b) w/o $\mathcal{L}_{pred}$  &      &  \cmark     &   \cmark    &    \xmark   &   \xmark    & 14.70 & 89.62 & 39.12  & 20.97 \\
    (c) w/o $\mathcal{L}_{act}$  &   \cmark    &     &   \cmark    &   34.86    &  49.13      & \xmark   & \xmark & 39.92 &   21.31    \\
    (d) w/o $\mathcal{L}_{rep}$  &    \cmark   &    \cmark    &      &    25.90   &  40.17    & 14.49 & 88.82 & \xmark     & \xmark   \\
    \midrule
    (e) only $\mathcal{L}_{pred}$ & \cmark &  &  &  33.23 & 47.28 & \xmark & \xmark & \xmark & \xmark \\
    (f) only $\mathcal{L}_{act}$ &  & \cmark &  & \xmark & \xmark & 14.32 & 91.13 & \xmark & \xmark \\
    (g) only $\mathcal{L}_{rep}$ &  &  &\cmark & \xmark & \xmark & \xmark & \xmark & 40.80 & 21.56 \\
    \midrule
    (h) Pixel-level latent &   \cmark   & \cmark     &       & 20.62 & 37.12 & \textbf{13.57} & \textbf{84.86} & 1.00     & 0.10   \\
    (i) Pixel-level latent &\cmark & \cmark & \cmark & 15.63  & 23.85 & 16.25  & 100.49 &  31.20 &  13.22  \\
    \bottomrule
    \end{tabular}%
    }
    \vspace{-1em}
  \label{tab:remove_one}%
\end{table*}%

{The TriFinger benchmark~\cite{wuthrich2020trifinger} involves a three-finger robot performing two tasks (reach cube and move cube) from an egocentric view. 
We freeze the pretrained models and use a 3-layer MLP as the policy network, training each task with 100 demonstrations following~\cite{majumdar2023we}.
As shown in Tab.~\ref{tab:trifinger}, EgoAgent achieves the highest success rates on both tasks, outperforming the egocentric representation models DoRA~\cite{venkataramanan2023imagenet} with increases of \textbf{+3.32\%} and \textbf{+3.9\%} respectively.
This result shows that by incorporating human action prediction into the learning process, EgoAgent demonstrates the ability to learn more effective representations that benefit both image classification and embodied manipulation tasks.
This highlights the potential of leveraging human-centric motion data to bridge the gap between visual understanding and actionable policy learning.}

\subsection{Ablation Study}

To evaluate the effectiveness of EgoAgent's specialized designs, we conduct several ablation studies using EgoAgent-300M with a short learning schedule of 14,400 iterations.

\noindent
{\textbf{Joint learning of representation, prediction, and action.}
EgoAgent demonstrates promising performance across the three tasks. To understand how these tasks interact during training, we conduct \textit{leave-one-out} experiments and \textit{one-task-only} experiments, as shown in Tab.~\ref{tab:remove_one} (a-g).}
We observe several key findings: (1) Removing any one task degrades the performance of the other two, highlighting the complementary roles of these tasks in learning.
Specifically,
removing the representation loss causes the largest drop in state prediction (\textbf{-11.87}\% in Top1 accuracy and \textbf{-13.55}\% in mAP), while removing the prediction loss significantly weakens representation learning, reducing Top1 accuracy by \textbf{-2.52}\% on ImageNet-100 and \textbf{-1.31}\% on ImageNet-1K.
This suggests that state prediction and visual representation are closely interconnected.
For motion prediction, excluding the prediction loss leads to the most significant decline, increasing MPJVE by \textbf{1.01}.
(2) Training on individual tasks yields worse performances on each task than learning all three tasks together, except for MPJPE in motion prediction.
We suspect that when trained only on motion prediction, the model focuses excessively on minimizing joint position errors. However, without context from the other tasks, it struggles to maintain smooth, consistent motion across frames, resulting in a higher MPJVE of \textbf{2.52}.

\noindent\textbf{Analysis of task dependencies.}
We also identify several interesting dependencies among the three tasks:
(1) \textit{Representation} serves as the foundation for both future state and action prediction as it captures the semantic information of the current world.
(2) \textit{Prediction and action} jointly provide effective supervision for representation learning by modeling world dynamics and predicting actions, improving representation by \textbf{1.16}\% in Top1 accuracy. However, training on prediction or action alone cannot enhance representation, as actions without accurate next-state predictions or next-state predictions conditioned on incorrect actions may provide noisy supervision.
(3) \textit{Representation and action} guide the model to integrate learned representations with action dynamics, leading to more accurate predictions of future states. Meanwhile, \textit{representation and prediction} work together to help the model understand how actions influence state transitions, refining its ability to generate actions.

\noindent{\textbf{Training in semantic feature space.} To evaluate the effectiveness of training agent models with a high-level semantic feature space, we use a pretrained VQGAN~\cite{esser2021taming}, which is designed to reconstruct pixel-level details of images, as the image tokenizer.
The results are shown in Tab.~\ref{tab:remove_one} (a,h,i).}
Training EgoAgent in VQGAN's pretrained pixel-level latent space results in significant performance drops in world state prediction and nearly nullifies the ability for visual representation. While motion prediction improves, likely due to the unified task formulation, the overall performance degradation makes this latent space unsuitable.
Forcing representation learning with this pixel-level latent space not only yields modest representation performance but also significantly weakens the other two tasks compared to learning in continuous semantic features. This suggests that high-level semantics are more effective than low-level details for understanding world states and distinguishing different contents in adjacent time steps.
\section{Conclusion and Future Work}
\label{sec:conclusion}
In this paper, we presented EgoAgent, a joint predictive agent model that simultaneously learns to represent world observations, predict future world states, and generate 3D human motions in egocentric worlds. EgoAgent integrates these three abilities within a joint embedding-action-prediction architecture with temporally asymmetric predictor-observer branches. 
Through extensive experiments, we demonstrated that EgoAgent outperforms existing state-of-the-art methods, with the joint learning of perception, prediction, and action tasks mutually reinforcing each other, leading to improved performance across all three capabilities.
Future work will explore incorporating finer-grained hand representations for precise tasks such as object manipulation and integrating long-term memory mechanisms for tasks requiring extended temporal dependencies.

\noindent\textbf{Acknowledgments:} This work was partially supported by the Zhejiang Provincial Natural Science Foundation of China (No. LR25F020003), NSFC (No. 62402427, No. U24B20154), Zhejiang University Education Foundation Qizhen Scholar Foundation, and Information Technology Center and State Key Lab of CAD\&CG, Zhejiang University. We also acknowledge the support from the JC STEM Lab of AI for Science and Engineering, funded by the Hong Kong Jockey Club Charities Trust, and the Research Grants Council of Hong Kong (No. CUHK14213224).
{
    \small
    \bibliographystyle{ieeenat_fullname}
    \bibliography{main}
}

\end{document}


\maketitlesupplementary

\section{Additional Results on Egocentric Future State Prediction}

\begin{figure}[b]
    \centering
    \includegraphics[width=\linewidth]{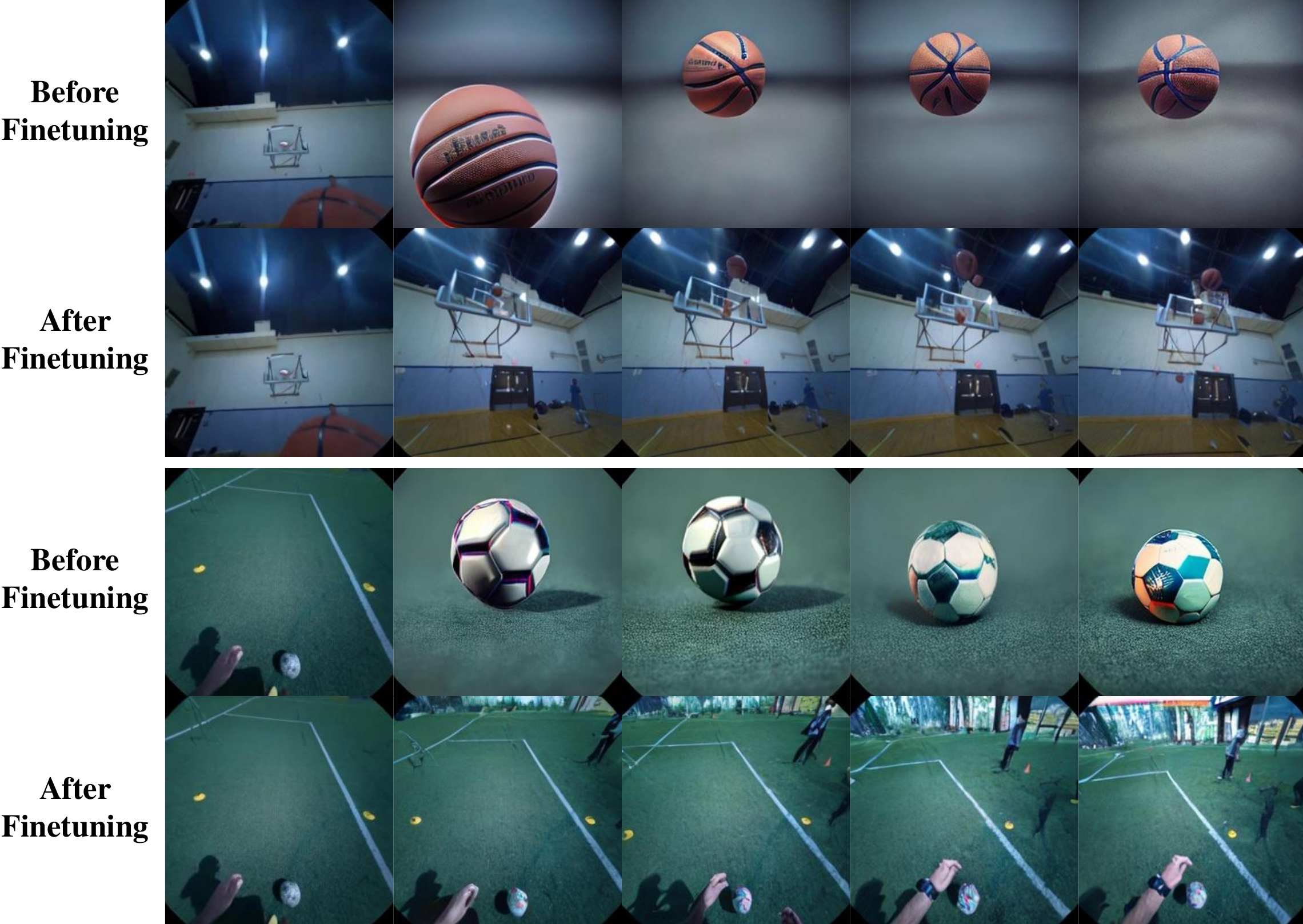}
    \caption{Comparison of OpenSora V1.1 first-frame-conditioned video generation results before and after finetuning on Ego-Exo4D. Fine-tuning enhances temporal consistency, but the predicted pixel-space future states still exhibit errors, such as inaccuracies in the basketball's trajectory.}
    \label{fig:opensora_finetune}
\end{figure}

In this section, we provide additional qualitative results on the egocentric future state prediction task. Additionally, we describe our approach to fine-tune a video diffusion model, OpenSora~\cite{opensora}, on the Ego-Exo4D dataset~\cite{grauman2024ego} and generate future video frames conditioned on initial frames as shown in Fig.~\ref{fig:opensora_finetune}.

\subsection{Visualizations and Comparisons}

\begin{figure*}
    \centering
    \includegraphics[width=\linewidth]{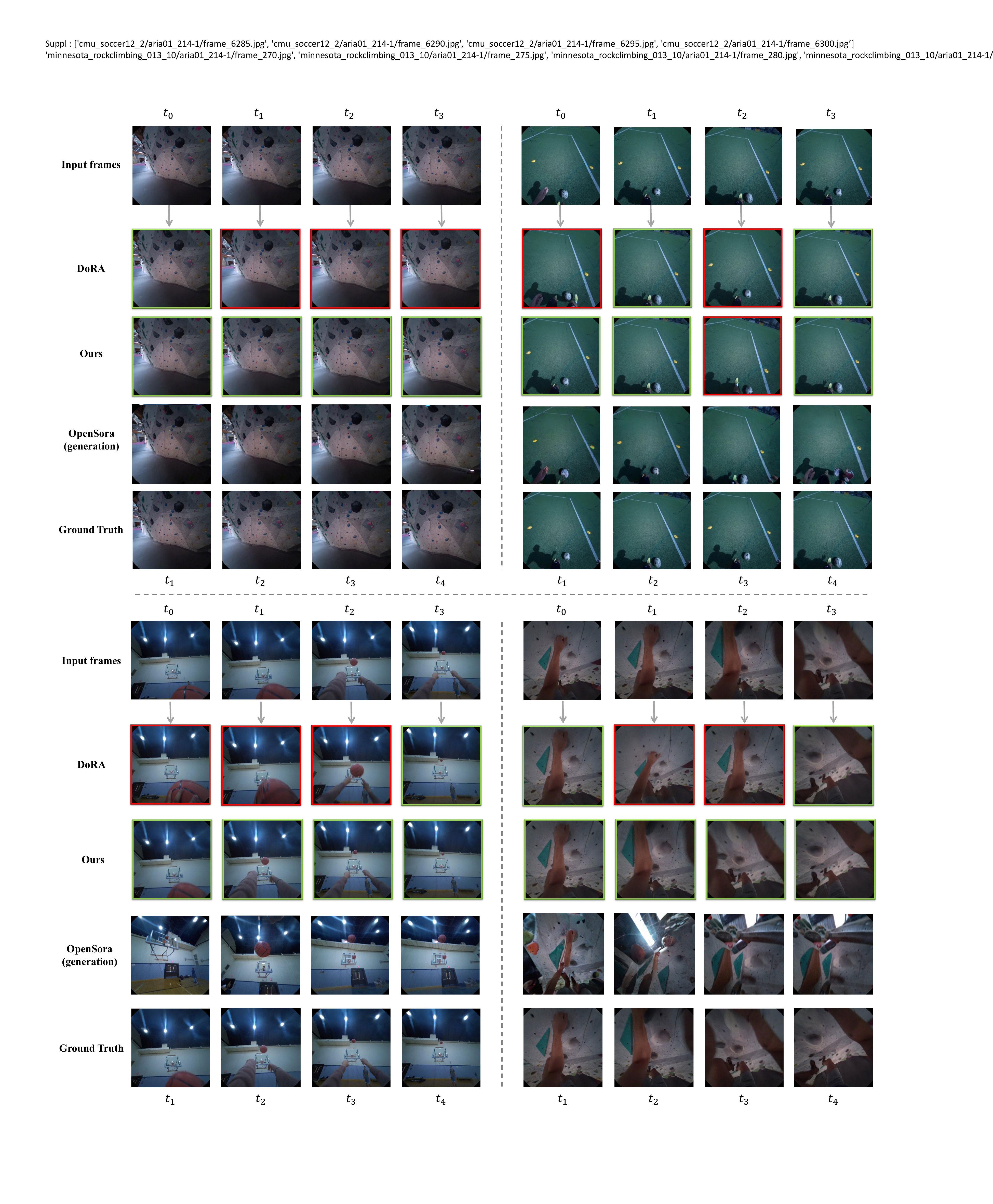}
    \caption{Retrieval and generation results for egocentric future state prediction. Correct and wrong retrieval images are marked with green and red borders, respectively.}
    \label{fig:supp pred}
\end{figure*}

We provide more visualizations of the prediction results from our EgoAgent, DoRA~\cite{venkataramanan2023imagenet}, and fine-tuned OpenSora~\cite{opensora} in different scenes in Fig.~\ref{fig:supp pred}. For OpenSora, when predicting the states of $t_k$, we use all the ground truth frames from $t_{0}$ to $t_{k-1}$ as conditions. As OpenSora takes only past observations as input and neglects human motion, it performs well only when the human has relatively small motions (see the top two cases in Fig.~\ref{fig:supp pred}), but can not adjust to large movements of the human body or quick viewpoint changes (see the bottom two cases in Fig.~\ref{fig:supp pred}).

\begin{figure*}[t]
    \centering
    \includegraphics[width=0.9\linewidth]{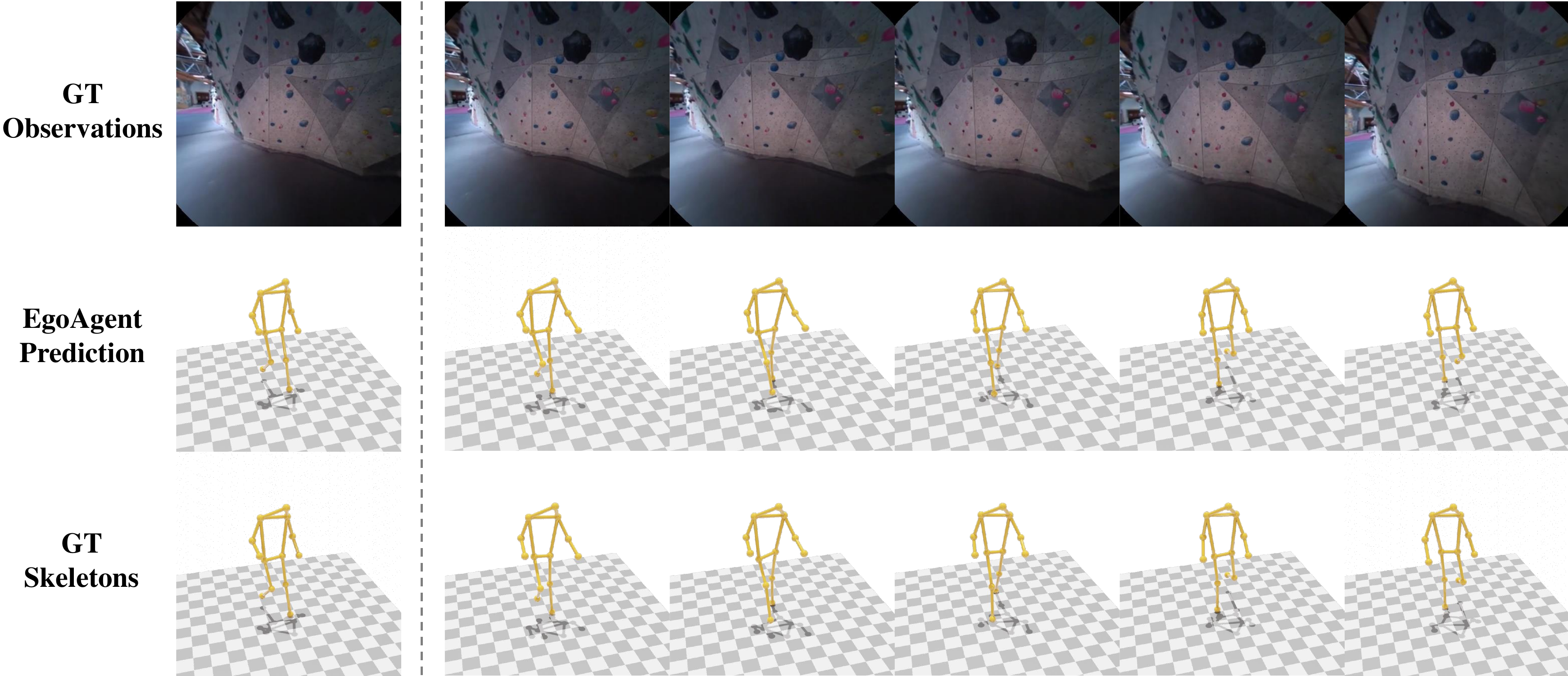}
    \vspace{-0.5mm}
    \caption{3D Human motion prediction results in scenes with minor changes in egocentric observations.}
    \vspace{-1.5mm}
    \label{fig:motion_prediction}
\end{figure*}

\subsection{Finetuning OpenSora on Ego-Exo4D}

OpenSora~\cite{opensora}, initially trained on Internet videos and images, produces severely inconsistent results when directly applied to infer future videos on the Ego-Exo4D dataset, as illustrated in Fig.~\ref{fig:opensora_finetune}.
To address the gap between general Internet content and egocentric video data, we fine-tuned the official OpenSora V1.1 checkpoint on the Ego-Exo4D training set for 50 epochs.
OpenSora V1.1 proposed a random mask strategy during training to enable video generation by image and video conditioning. We adopted the default masking rate, which applies: 75\% with no masking, 2.5\% with random masking of 1 frame to 1/4 of the total frames, 2.5\% with masking at either the beginning or the end for 1 frame to 1/4 of the total frames, and 5\% with random masking spanning 1 frame to 1/4 of the total frames at both the beginning and the end.

As shown in Fig.~\ref{fig:opensora_finetune}, despite being trained on a large dataset, OpenSora struggles to generalize to the Ego-Exo4D dataset, producing future video frames with minimal consistency relative to the conditioning frame. While fine-tuning improves temporal consistency, the moving trajectories of objects like the basketball and soccer ball still deviate from real physical laws. Compared with our feature space prediction results, this suggests that training world models in a reconstructive latent space is more challenging than training them in a feature space.

\section{Additional Results on 3D Human Motion Prediction}

We present additional qualitative results for the 3D human motion prediction task, highlighting a particularly challenging scenario where egocentric observations exhibit minimal variation. This scenario poses significant difficulties for video-conditioned motion prediction, as the model must effectively capture and interpret subtle changes. As demonstrated in Fig.~\ref{fig:motion_prediction}, EgoAgent successfully generates accurate predictions that closely align with the ground truth motion, showcasing its ability to handle fine-grained temporal dynamics and nuanced contextual cues.

\section{Action Affects the Perceptual Area}

\begin{figure}[t]
    \centering
    \includegraphics[width=0.7\linewidth]{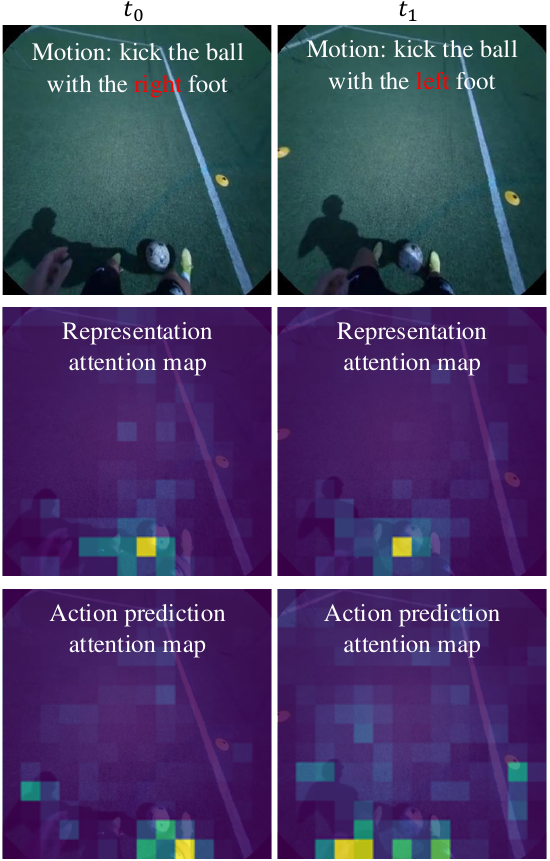}
    \vspace{-0.5mm}
    \caption{Attention map of EgoAgent performing visual representation and action prediction task. EgoAgent attends to the moving body part in the image when predicting future actions.}
    \vspace{-1.5mm}
    \label{fig:attention}
\end{figure}

We visualize the attention map of EgoAgent's first transformer head in the representation and action prediction tasks.
As shown in Fig.~\ref{fig:attention}, when provided with only the egocentric image, EgoAgent focuses on the ball. Upon receiving the action query token, it shifts attention to the human body part (the right foot at $t_0$ and the left foot at $t_1$) in line with the ground truth body motion, indicating that actions can guide the model to focus on task-related areas.

\section{OpenSora for Image Classification}

In this section, we detail the process of extracting features from OpenSora V1.1~\cite{opensora} (without fine-tuning) for an image classification task. Following the approach of~\cite{xiang2023denoising}, we leverage the insight that diffusion models can be interpreted as multi-level denoising autoencoders. These models inherently learn linearly separable representations within their intermediate layers, without relying on auxiliary encoders. The quality of the extracted features depends on both the layer depth and the noise level applied during extraction.

\begin{table}[t]
\centering
\caption{$k$-NN evaluation results of OpenSora V1.1 features from different layer depths and noising scales on ImageNet-100. Top1 and Top5 accuracy (\%) are reported.}
\label{tab:opensora-knn}
\resizebox{0.95\linewidth}{!}{%
\begin{tabular}{lcccccc}
\toprule
\multirow{2}{*}{Timesteps} & \multicolumn{2}{c}{First Layer} & \multicolumn{2}{c}{Middle Layer} & \multicolumn{2}{c}{Last Layer} \\
\cmidrule(r){2-3}   \cmidrule(r){4-5}  \cmidrule(r){6-7}  & Top1           & Top5           & Top1            & Top5           & Top1           & Top5          \\
\midrule
32        &  6.10           & 18.20             & 34.04               & 59.50             & 30.40             & 55.74             \\
64        & 6.12              & 18.48              & 36.04               & 61.84              & 31.80         & 57.06         \\
128       & 5.84             & 18.14             & 38.08               & 64.16              & 33.44       & 58.42 \\
256       & 5.60             & 16.58              & 30.34               & 56.38              &28.14          & 52.32        \\
512       & 3.66              & 11.70            & 6.24              & 17.62              & 7.24              & 19.44  \\          
\bottomrule
\end{tabular}%
}
\end{table}

As shown in Tab.~\ref{tab:opensora-knn}, we first evaluate $k$-NN classification performance on the ImageNet-100 dataset using three intermediate layers and five different noise scales. We find that a noise timestep of 128 yields the best results, with the middle and last layers performing significantly better than the first layer.
We then test this optimal configuration on ImageNet-1K and find that the last layer with 128 noising timesteps achieves the best classification accuracy.

\section{Data Preprocess}
For egocentric video sequences, we utilize videos from the Ego-Exo4D~\cite{grauman2024ego} and WalkingTours (WT)~\cite{venkataramanan2023imagenet} datasets.
The original resolution of Ego-Exo4D videos is 1408$\times$1408, captured at 30 fps. We sample one frame every five frames and use the original resolution to crop local views (224$\times$224) for computing the self-supervised representation loss~\cite{caron2021emerging}. For computing the prediction and action loss, the videos are downsampled to 224$\times$224 resolution.
WT primarily consists of 4K videos (3840$\times$2160) recorded at 60 or 30 fps. Similar to Ego-Exo4D, we use the original resolution and downsample the frame rate to 6 fps for representation loss computation.
As Ego-Exo4D employs fisheye cameras, we undistort the images to a pinhole camera model using the official \emph{Project Aria Tools} to align them with the WT videos.

For motion sequences, the Ego-Exo4D dataset provides synchronized 3D motion annotations and camera extrinsic parameters for various tasks and scenes. While some annotations are manually labeled, others are automatically generated using 3D motion estimation algorithms from multiple exocentric views. To maximize data utility and maintain high-quality annotations, manual labels are prioritized wherever available, and automated annotations are used only when manual labels are absent.
Each pose is converted into the egocentric camera's coordinate system using transformation matrix derived from the camera extrinsics. These transformation matrices also enable the computation of trajectory vectors for each frame in a sequence. Beyond the $x, y, z$ coordinates, a visibility dimension is appended to account for keypoints invisible to all exocentric views. Finally, a sliding window approach segments sequences into fixed-size windows to serve as input to the model. Note that we do not downsample the frame rate of 3D motions.

\section{Training Details}
\subsection{Architecture Configurations}
In Tab.~\ref{tab:arch}, we provide detailed architecture configurations for EgoAgent following the scaling-up strategy of InternLM~\cite{team2023internlm}. To maintain the scaling-up and generalization ability, we do not modify the internal modules in InternLM, \emph{i.e.}, we adopt the RMSNorm~\cite{zhang2019root} and 1D RoPE~\cite{su2024roformer}. We show that, without specific modules designed for vision tasks, EgoAgent can perform well on egocentric vision and action tasks.

\begin{table}[t]
  \centering
  \caption{Architecture configurations of EgoAgent.}
  \resizebox{0.8\linewidth}{!}{%
    \begin{tabular}{lcc}
    \toprule
          & EgoAgent-300M & EgoAgent-1B \\
          \midrule
    Depth & 22    & 22 \\
    Embedding dim & 1024  & 2048 \\
    Number of heads & 8     & 16 \\
    MLP ratio &    8/3   & 8/3 \\
    $\#$Param.  & 284M & 1.13B \\
    \bottomrule
    \end{tabular}%
    }
  \label{tab:arch}%
\end{table}%

Tab.~\ref{tab:io_structure} presents the detailed configuration of the embedding and prediction modules in EgoAgent, including the image projector ($\text{Proj}_i$), representation head/state prediction head ($\text{MLP}_i$), action projector ($\text{Proj}_a$) and action prediction head ($\text{MLP}_a$).
Note that the representation head and the state prediction head share the same architecture but have distinct weights.

\begin{table}[t]
\centering
\caption{Architecture of the embedding ($\text{Proj}_i$, $\text{Proj}_a$) and prediction ($\text{MLP}_i$, $\text{MLP}_a$) modules in EgoAgent. For details on module connections and functions, please refer to Fig.~2 in the main paper.}
\label{tab:io_structure}
\resizebox{\linewidth}{!}{%
\begin{tabular}{lcl}
\toprule
       & \multicolumn{1}{c}{Norm \& Activation} & \multicolumn{1}{c}{Output Shape}  \\
\midrule
\multicolumn{3}{l}{$\text{Proj}_i$ (\textit{Image projector})} \\
\midrule
Input image  & -          & 3$\times$224$\times$224 \\
Conv 2D (16$\times$16) & -       & Embedding dim$\times$14$\times$14    \\
\midrule
\multicolumn{3}{l}{$\text{MLP}_i$ (\textit{State prediction head} \& \textit{Representation head)}} \\
\midrule
Input embedding  & -          & Embedding dim \\
Linear & GELU       & 2048          \\
Linear & GELU       & 2048          \\
Linear & -          & 256           \\
Linear & -          & 65536     \\
\midrule
\multicolumn{3}{l}{$\text{Proj}_a$ (\textit{Action projector})} \\
\midrule
Input pose sequence  & -          & 4$\times$5$\times$17 \\
Conv 2D (5$\times$17) & LN, GELU   & Embedding dim$\times$1$\times$1    \\
\midrule
\multicolumn{3}{l}{$\text{MLP}_a$ (\textit{Action prediction head})} \\
\midrule
Input embedding  & -          & Embedding dim$\times$1$\times$1 \\
Linear & -          & 4$\times$5$\times$17     \\
\bottomrule
\end{tabular}%
}
\end{table}

\subsection{Training Configurations}
In Tab.~\ref{tab:training hyper}, we provide the detailed training hyper-parameters for experiments in the main manuscripts.
The training uses the AdamW optimizer~\cite{loshchilov2017decoupled} with momentum parameters $\beta_1=0.9$ and $\beta_2=0.999$. The base learning rate is set at $6\times 10^{-4}$, with a cosine learning rate schedule and a base weight decay of $0.04$, transitioning to an end weight decay of $0.4$. A batch size of $1920$ is employed for $72,000$ iterations, with $1,800$ warm-up iterations using a linear schedule. Gradient clipping is applied at a value of $1.0$, and data is processed in Float-16 precision. Additionally, the exponential moving average (EMA) momentum is set to $0.996$. The normalization epsilon is fixed at $1\times 10^{-6}$ for stability in training.

\begin{table}[t]
  \centering
  \caption{Hyper-parameters for training EgoAgent.}
  \resizebox{0.86\linewidth}{!}{%
    \begin{tabular}{lc}
    \toprule
    Training Configuration & EgoAgent-300M/1B \\
    \midrule
    Training recipe: &  \\
    optimizer & AdamW~\cite{loshchilov2017decoupled} \\
    optimizer momentum & $\beta_1=0.9, \beta_2=0.999$ \\
    \midrule
    Learning hyper-parameters: &  \\
    base learning rate & 6.0E-04 \\
    learning rate schedule & cosine \\
    base weight decay & 0.04 \\
    end weight decay & 0.4 \\
    batch size & 1920 \\
    training iterations & 72,000 \\
    learning rate warm-up iterations & 1,800 \\
    warm-up schedule & linear \\
    gradient clip & 1.0 \\
    data type & Float-16 \\
    norm epsilon & 1.0E-06 \\
    \midrule
    EMA hyper-parameters: &  \\
    momentum & 0.996 \\
    \bottomrule
    \end{tabular}%
    }
  \label{tab:training hyper}%
\end{table}%

\clearpage

{
    \small
    \bibliographystyle{ieeenat_fullname}
    \bibliography{main}
}